\documentclass[draftcls,12pt,onecolumn]{IEEEtran} 
\usepackage{amsmath,epsfig,hyperref}
\usepackage{setspace}
\usepackage{array}
\usepackage{graphicx}
\usepackage{listings}
\usepackage{verbatim}
\usepackage[hang,raggedright,scriptsize]{subfigure}
\usepackage{calc}
\usepackage{amsmath}
\usepackage{amsfonts}
\usepackage{amssymb}
\usepackage{amsbsy}
\usepackage{amstext}
\usepackage{amsthm}
\usepackage{multicol}
\usepackage{multirow}
\usepackage[ansinew]{inputenc}
\usepackage{cite}
\usepackage{color}
\usepackage{booktabs}
\usepackage{lscape}
\usepackage{wrapfig}
\usepackage{slashbox}
\usepackage{ textcomp }

\newcommand{\blu}[1]{\textcolor[rgb]{0,0,0}{#1}}
\newcommand{\blue}[1]{\textcolor[rgb]{0,0,0}{#1}}

\newcommand{\x}{\mathbf{x}}

\newcommand{\mytitle}{Recent Advances in Domain Adaptation for the Classification of Remote Sensing Data}

\begin{document}

\title{\mytitle}

\author{Devis Tuia,~\IEEEmembership{Senior Member,~IEEE}, Claudio Persello,~\IEEEmembership{Member,~IEEE}, Lorenzo Bruzzone,~\IEEEmembership{Fellow,~IEEE}
\thanks{Manuscript received 2015;}

\thanks{
D. Tuia is with the Department of Geography, University of Zurich, 8057 Zurich, Switzerland. Email: \emph{devis.tuia@geo.uzh.ch},\newline
C. Persello is with the faculty of Geo-Information Science and Earth Observation (ITC), University of Twente, The Netherlands. Email \emph{c.persello@utwente.nl}, \newline
L. Bruzzone is with the Dipartimento di Ingegneria e Scienza dell'Informazione, University of Trento, Italy. Email: \emph{lorenzo.bruzzone@disi.unitn.it}}
}

\markboth{IEEE Geoscience and Remote Sensing Magazine 2016, Preprint, full version: 10.1109/MGRS.2016.2548504}{Title: \mytitle}
\maketitle

\begin{abstract}
\textbf{This is the pre-acceptance version, to read the final version published in 2016 in the IEEE Geoscience and Remote Sensing Magazine, please go to: \href{https://doi.org/10.1109/MGRS.2016.2548504}{10.1109/MGRS.2016.2548504}}\\
\blu{
The success of supervised classification of remotely sensed images acquired over large geographical areas or at short time intervals strongly depends on the representativity of the samples used to train the classification algorithm and to define the model. When training samples are collected from an image (or a spatial region) different from the one used for mapping, spectral shifts between the two distributions are likely to make the model fail. Such shifts are generally due to differences in acquisition and atmospheric conditions or to changes in the nature of the object observed. 
In order to design classification methods that are robust to data-set shifts, recent remote sensing literature has considered solutions based on \emph{domain adaptation} (DA) approaches.
Inspired by machine learning literature, 
several DA methods have been proposed  to solve specific problems in remote sensing data classification. This paper provides a critical review of the recent advances in DA for remote sensing and presents an overview of methods divided into four categories: i) invariant feature selection; ii) representation matching; iii) adaptation of classifiers and iv) selective sampling. We provide an overview of recent methodologies, as well as examples of application of the considered techniques to real remote sensing images characterized by very high spatial and spectral resolution. Finally, we propose guidelines to the selection of the method to use in real application scenarios. 
}
\end{abstract}

\section{Introduction} \label{sec:Intro} 

{
With the advent of the new generation of satellite missions, which are often made up of constellations of satellites with short revisit time and very high resolution sensors, the amount of  remote sensing images available has increased significantly. Nowadays, monitoring of dynamic processes has become possible~\cite{deJ11,Pac14}, as well as the use of several data sources to address bio-physical parameter estimation and classification problems~\cite{Wal12,Lia14,Amoros-Lopez13,Wal14}. As a consequence, analysts have the opportunity to use} multitemporal and multisource images for tasks such as repetitive monitoring of the territory, change detection, image mosaicking and large scale processing (i.e. processing involving many image tiles)~\cite{Gom14}. 

\begin{figure}[!t]
\centering
\includegraphics[width=.9\linewidth]{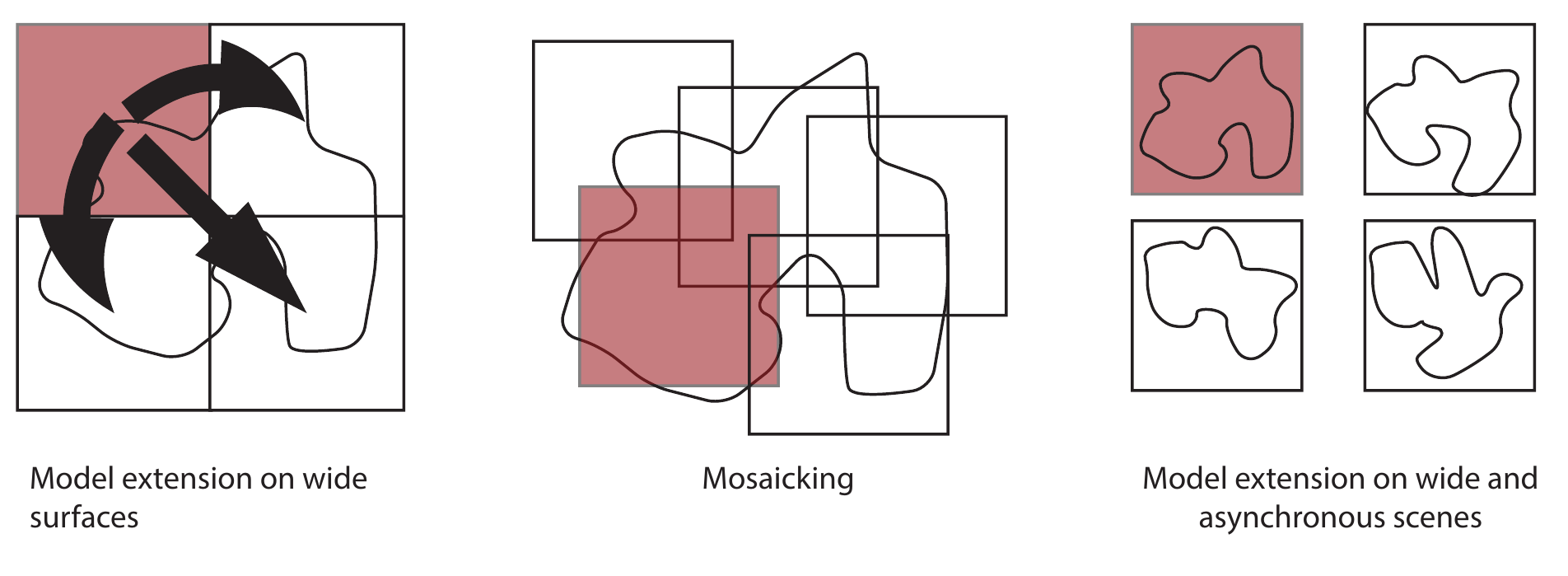}
\caption{Examples of cases where domain adaptation is necessary to extend a model to new image acquisitions. In all three cases, the images can be of different sensors and only the image in red has extensive reference labels (i.e. can be used for training an accurate supervised model).}
\label{fig:probs}
\end{figure}

Remote sensing is therefore facing new opportunities. However, such opportunities cannot be seized, unless they come with the capability to provide accurate products in a timely manner. A bottleneck of supervised image processing-based pipelines is the need of training \blue{the model on reference points that are specific to every acquisition}: to be accurate, most models need to be trained on known samples coming from the image under study. Since obtaining new ground samples of high quality for each image acquisition is not realistic, to retrain/adapt an existing model without such ground samples becomes mandatory. Figure~\ref{fig:probs} illustrates situations where adaptive models might be of great use: in these situations (which correspond to those considered in this paper) only one image -- the one in red in the figure -- has sufficient reference labels (e.g. obtained in an extensive ground campaign) whereas the others have no labeled samples or have them only in an insufficient number. This setting is more realistic, since, on the one hand, extensive labeling cannot follow the pace of image acquisitions and, on the other hand, repetitive ground campaigns are often simply not an option, mainly for economic and manpower reasons. Indeed, gathering ground information is costly and cannot always be performed by photointerpretation: this is particularly true when the task concerns very large areas or considers quantities that cannot be photointerpreted by an analyst, such as chlorophyll concentrations~\cite{Verrelst2013}, plant water stress~\cite{Beh14} or tree species~\cite{Dal12}.

A possible solution would be to bypass the problem and assume that the model already available is robust enough to process the new \blu{images} accurately. Despite the fact that this is only possible in cases where the new image is acquired by the same sensor as the previous one, it is well known that the direct application of a pre-trained model on a new dataset often provides  poor results. This is because the spectra observed in the new scene, even though representing the same types of objects, are different from those of the scene used for training. \blu{The differences} can be related to a series of deformations, or \emph{shifts}, related to a variety of effects such as a biased sampling in the spatial domain (typically if the ground sampling has been focused on a region non-representative of the new scene),  changes in the acquisition conditions (including illumination or acquisition angle), or seasonal changes.
When the new data are acquired by a different sensor, the strategy above is simply not applicable, as most models require that all images (or domains) provide samples of the same dimensionality (and where each dimension has the same meaning) at test time. In this case, fusion strategies exist, but generally apply only to certain combination of sensors and use only the bands that these sensors have in common (or that are reasonably similar)~\cite{Wal12}. This prevents to reuse models \blu{already} trained on the first image \blu{of the region that becomes} available (which can be crucial, for example, in post-catastrophe interventions) and to exploit sensors synergies in multi-sensorial schemes.

From these examples it is clear that, to process efficiently and accurately remote sensing images, modern processing systems must be designed to be robust to changes in acquisition conditions, temporal shifts and (ideally) be adaptive to sensors differences. The need for adapting existing models has been acknowledged since many years, as shown by the signature extension community~\cite{Fle75,Olt05}, but the change in the amounts and nature  of data -- as well as their resolutions -- created the need for a new research direction. In this paper, we advocate that the solution can be found in \emph{domain adaptation} (DA) strategies, a field deeply rooted in statistical and machine learning~\cite{Qui09,Pan09}.

Domain adaptation basically aims at adapting models trained to solve a specific task to a new-but-yet-related task, for which the knowledge of the initial model is sufficient, although not perfect. As a traditional example in computer vision, DA method have been deployed to adapt classifiers recognizing objects in pictures  from commercial websites to the recognition of objects photographed form simple webcameras~\cite{saenko10}. In this example, the classifier is presented a problem with the same objective (classifying pictures into a limited set of object classes) and the same features, but where the data relations are slightly different (for example, in Amazon.com the pictures have no background and the object is mostly in the centre of the image, while in the \blu{case of the} webcam \blu{images} this is not the case). Domain adaptation is therefore used to adapt the classifier that is accurate on Amazon.com to the new data distribution. Of course, this is just one example: in the domain adaptation literature, models are modified to adapt to new data spaces (multimodal), related tasks (multitask) or subtle changes in probability distributions (see a recent review in~\cite{Pat15}). The connections to multi-temporal, multi-sensor and multi-resolution image classification tasks above are strong~\cite{Gom14}.  

The aim of this review paper is to provide an introduction to the domain adaptation field and to provide  examples of application of DA techniques in remote sensing. With this in mind, we draw a taxonomy of the DA strategies that have been proposed in recent remote sensing literature \blu{and discuss} their strengths and weaknesses. We also provide a series of practical examples about the use of DA in high to very-high resolution image processing tasks. We will not enter into the technical details of  specific DA literature: for readers interested in these, we refer to the recent surveys published in~\cite{Qui09,Pan09,Sug12,Pat15}.

The remainder of the paper is organized as follows: Section~\ref{sec:DA} is an overview of the DA problem and fixes the terminology and notations that will be used throughout the paper. Section~\ref{sec:Tax} draws a taxonomy of DA approaches in four families, respectively based on i) invariant feature selection; ii)  matching of data representation; iii)  adaptation of the classifier used  and iv) limited, but effective sampling in the new domain(s). These families are described together with recent references and key examples of remote sensing applications. Section~\ref{sec:concl} concludes the paper.

\section{Transfer Learning and Domain Adaptation} \label{sec:DA}

Transfer learning problems arise when inference has to be made on processes which are not stationary over time or space. As we have discussed in the introduction, this is the case in the analysis of remote sensing images, where different acquisitions are typically subject to different conditions (e.g., illumination, viewing angle, soil moisture, topography). Such differences affect the observed spectral signatures of the land-cover types and therefore the distribution of the information classes in the feature space~\cite{Mat14b}. 
Different transfer learning problems (and techniques to tackle them) have been considered in the literature: domain adaptation, multi-task learning, 
domain generalization, sample selection bias, covariate shift \cite{Pan09}.
In this paper we will focus on DA, which is a particular form of transfer learning.

\begin{figure}[t]
\centerline{\includegraphics[width=0.7\columnwidth]{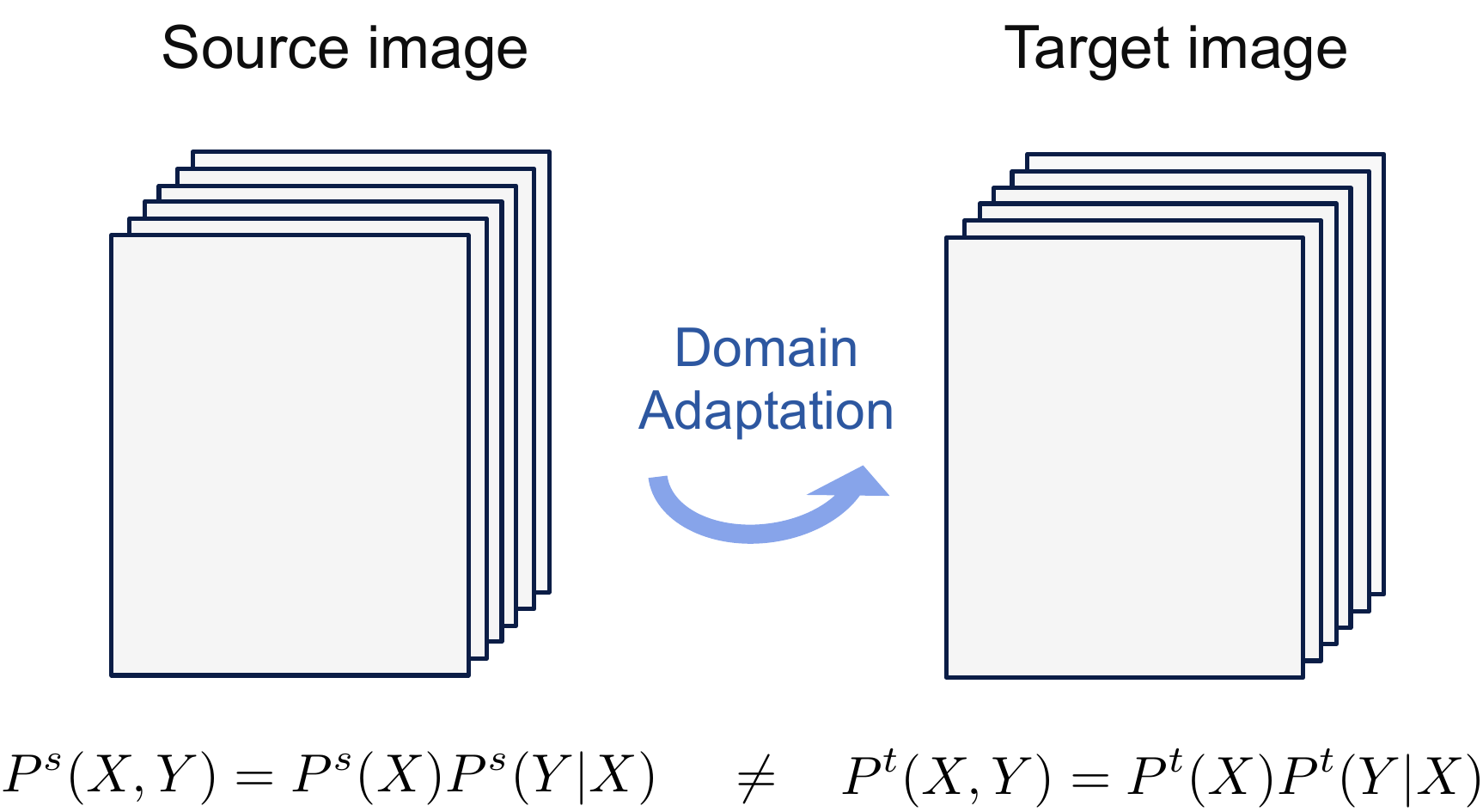}}
\caption{Graphical representation of the domain adaptation problem in the context of remote sensing image classification.  Source and target images can be acquired on different geographical areas (but with similar land cover characteristics) or on the same area at different time. The two images are associated with two different joint distributions, which characterize the two classification problems. The two distributions can differ because of different acquisition conditions, e.g., illumination, viewing angle, soil moisture, topography.
 }
 \label{fig:DA}
 \end{figure}

Let us consider two domains, called {\it source} and {\it target domain}, associated with two images acquired on different geographical areas (but with similar land-cover characteristics) or on the same area at different time {instants}. 
Figure \ref{fig:DA} \blu{sketches} the DA problem in the context of remote sensing image classification.
The source and target domains are associated with the joint probability distributions $P^s(X,Y)$ and $P^t(X,Y)$, respectively. 
The two joint probabilities define the classification problems on the two domains, where $X$ is the input (vector) variable (spectral bands of the source image with possible additional features used to characterize the contextual information of the single pixel) and $Y$ is the output variable associated with \blu{a set of classes} (land-cover or land-use information). The aim of DA methods is to adapt a classifier trained on the source domain to make predictions on the target domain. 

{\it Supervised DA} assumes that labeled samples are available for both domains. The labeled sets ${T^s=\{ (\x_1,y_1), \dots , (\x_n, y_n)\}}$ and ${T^t=\{ (\x_1,y_1), \dots , (\x_m, y_m)\}}$ are the source and the target-domain training sets, respectively. {Supervised} DA methods focus on challenging situations where {labeled} target-domain samples are less numerous than {those available in the source domain}, i.e., ${m<<n}$. In such conditions, the proper usage of source-domain information is very important to solve the target problem.
Most of the works in DA assume that source and target domains share the same set of classes. Only few papers address DA considering differences in the set of classes between source and target \cite{Tui11d,Bah12, Dem13,Jun13}.

{\it Semisupervised DA} methods assume that a training set is available only for the source domain, whereas target-domain information is limited to a set of unlabeled samples $U^t=\{ \x_1, \dots , \x_m \}$. 
This  DA setting is more \blu{challenging} than the supervised case and requires important assumptions on the relationship between source and target domains to make the algorithm converge to a consistent solution on the target domain.
All DA methods are based on the assumption that $P^s(X,Y)$ and $P^t(X,Y)$ are different, but close enough to ensure that the  source-domain information can be of help for solving the target-domain learning problem. 
On the one hand, if source and target domain are arbitrarily different, there is no hope that the source-domain information will provide an advantage in solving the task in the target domain. On the other hand, if $P^s(X,Y) = P^t(X,Y)$, no adaptation is necessary and the model trained on the source can be readily applied to the target. Semisupervised DA methods are effective in situations that lie in between these two extreme cases. 

{\it Unsupervised DA} methods are the last family, which assumes that two unlabeled domains have to be matched. This is the most difficult case, because label information is not available for any domain. In this situation, DA methods aim at matching the marginal distributions of the two domains $P^s(X)$ and $P^t(X)$ without knowledge on the learning task (classification or regression).
Unsupervised methods can be used as preprocessing of any analysis task (clustering, density estimation), but they imperatively need to have datasets with similar structural properties before adaptation. 
Unsupervised DA models are generally feature extractors or matching algorithms that exploit the geometrical structure of the data. 

Related problems 
are {\it sample selection bias} and {\it covariate shift} \cite{Heckman1979, Zadrozny2004, Huang07}. Sample selection bias originates when the available training samples are not independently and randomly selected from the underlying distribution (i.e., the training set in not a random sample of the population). This situation is very likely to occur in remote sensing problems where training points are usually selected and labeled through photo-interpretation or field surveys.
Covariate shift is a particular case of sample selection bias where the bias depends only on the input variable $X$ (and not on $Y$).
Different operational conditions that result in biased training samples are discussed in \cite{Persello2014b}.
Clearly, a training set ${T=\{ (\x_1,y_1), \dots , (\x_n, y_n)\}}$ obtained under sample selection bias leads to skewed estimation of the true underlying distribution of the classes, resulting in an estimated $\hat{P}(X,Y) \neq P(X,Y)$. For this reason, the effect of a sample selection bias is similar to the DA problem described above. The covariate shift problem is generally not as severe as the general sample selection bias and the DA problem. 
Let us consider that both training sets on the source and target are samples with bias (bias depending on X only, i.e., covariate shift) from the same joint distribution. The joint probabilities on the two domains can be factorized as follows: $P^s(X,Y)=P^s(X)P^s(Y|X)$ and $P^t(X,Y)=P^t(X)P^t(Y|X)$. In this particular case of covariate shift, the estimated conditional probabilities will be approximately equal, while the marginal distributions will in general be different, i.e., $\hat{P}^s(Y|X)\approx \hat{P}^t(Y|X)$ and $\hat{P}^s(X) \neq \hat{P}^t(X)$. 

\begin{figure}[t]
\centerline{\includegraphics[width=0.7\columnwidth]{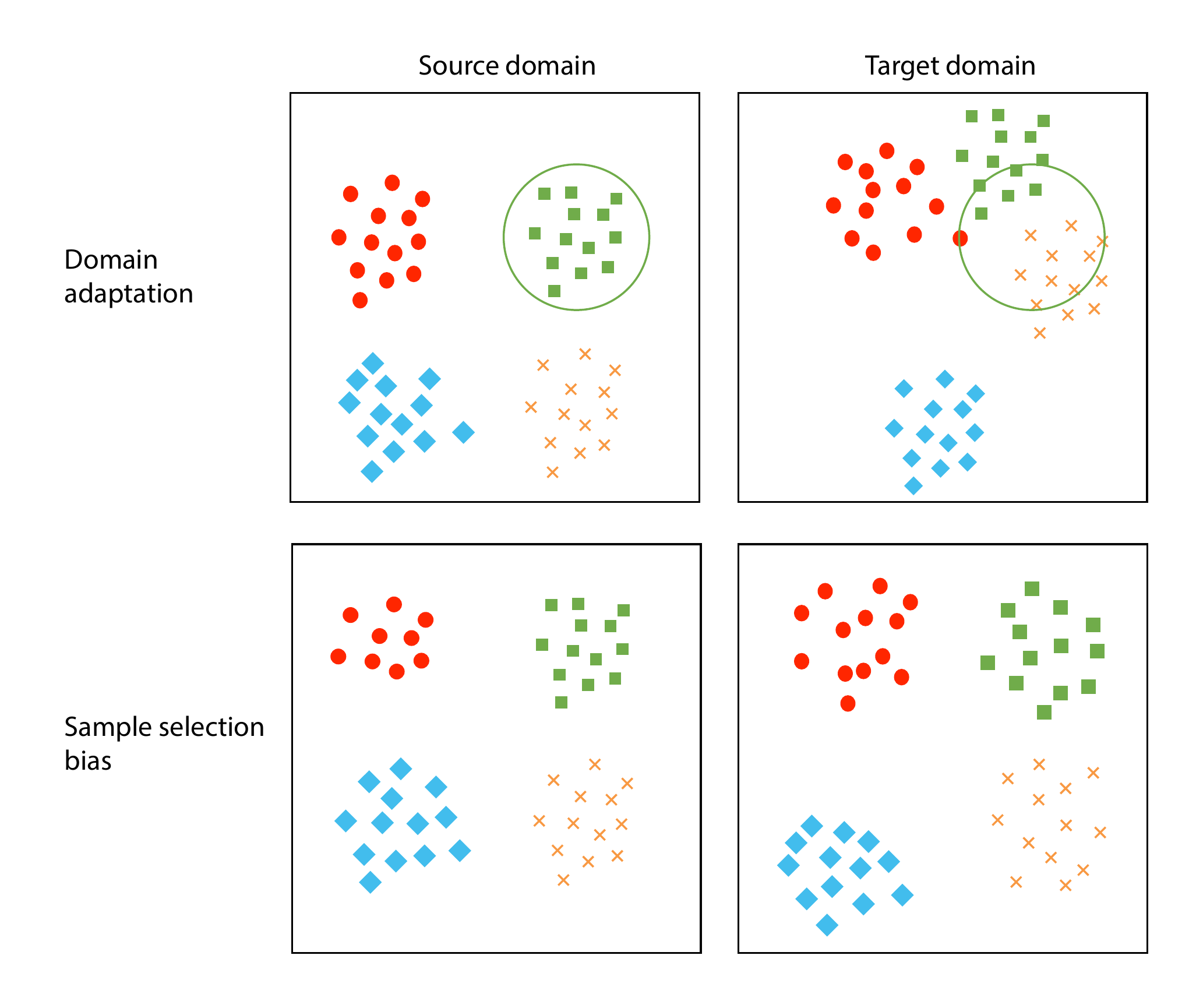}}
\caption{Explanatory examples of domain adaptation and sample selection bias problems. The plots represent labeled samples in a bi-dimensional feature space on the source and target domain. A four-classes classification problem is considered. In the case of DA, the class-conditional densities may vary from the source to the target domain, resulting in a significantly different classification problem \blu{(the green circle corresponds to the location of the green class in the source domain)}. In the case of sample selection bias, the aforementioned issue will not occur resulting in a milder type of shift from the source domain.
}
 \label{fig:DAvsSSB}
 \end{figure}

Figure~\ref{fig:DAvsSSB} illustrates the two different issues of DA and sample selection bias with two explanatory examples. The plots report the distributions of the labeled samples from two domains in a bi-dimensional feature space. A four-classes classification problem is considered. In the case of DA, the class-conditional densities may change from the source to the target domain, (possibly) resulting in a significantly different classification problem. Note that the distribution of the classes in the target domain may overlap with different classes on the source domain \blu{(see the green circle in the target domain, representing the location of the green class in the source)}. \label{sec:classOverlap} In this situation, training samples of the source domain can be misleading for the classification of the target domain~\cite{Per12, Persello2013}.
In the case of sample selection bias, source and target-domain samples are drawn (with bias) from the same underlying distribution, resulting in general in a milder type of shift with respect to DA (the problem of cross-domain class overlap is not likely to occur in sample selection bias problems).

In real RS problems, it is expected that the two aforementioned issues 
may occur at the same {time} and can be profoundly entangled. This means that: 1) the two (theoretical) underlying distributions ${P^s(X,Y)}$ and ${P^t(X,Y)}$ associated with source and target domains, respectively, may differ because of changes in the image acquisition conditions (e.g., radiometric conditions), and 2) the available training samples could be not fully representative of the statistical populations in their respective domains and will therefore lead to biased estimations of the class probabilities and to inaccurate classification models. 
In remote sensing literature, several techniques have been presented aiming at solving the transfer learning problem irrespectively of the cause of the data-set shift between source and target domains. The next section will present the different families of techniques that have been proposed in remote sensing image processing.

\section{A taxonomy of adaptation methods} \label{sec:Tax}

Adapting a model trained on one image to another (or a series of \blu{new images}) can be performed in different ways. In this section, we detail recent approaches proposed in the remote sensing literature by grouping them in four categories:

\begin{itemize}

\item \blu{\emph{Selection of invariant features}, where a set of the input features (original bands or additional features extracted from the remote sensing image) which are not affected by the shifting factors are identified and selected before training the classification algorithm. Accordingly, the features affected by the most severe data-set shift are removed and the classifier considers a feature space showing higher stability across domains.
An alternative approach is to encode invariance by including additional synthetic labeled samples in the training set in order to extract features that better model the intra-class variability across the domains.}
\item \emph{Adaptation of the data distributions}, where the data distributions of the target and source domains are made as similar as possible in order to keep the classifier unchanged. With respect to the previous family, these methods work on the original input spaces and try to extract a common space, where all domains can be treated equally. This is generally {achieved by means of joint feature extraction.}

\item \blu{\emph{Adaptation of the classifier}, where the classification model defined by training on the source domain is adapted to the target domain by considering unlabeled samples of the target domain. In this case, the data distributions remain unchanged and the classifier is adapted to the target data distribution \blu{using strategies based on semi-supervised learning}. 
\item \emph{Adaptation of the classifier by active learning}}, where the adaptation is performed by providing a limited amount of well-chosen labeled samples from the target domain. This is a special case of the \blu{previous} family, where we allow some new labeled examples to be sampled in the target domain in order to retrain the model iteratively. Due to their acquisition cost, these samples need to be selected well according to their potential to lead the model towards the desired target classifier.
\end{itemize}

The rest of this section details recent advances for these four families. We will limit the discussion to approaches specific to remote sensing literature and invite the interested reader to consult the specialized machine learning and computer vision  literature in~ \cite{Pan09,Qui09,Pat15}.

\newcommand{\argmin}{\operatornamewithlimits{argmin}}
\newcommand{\argmax}{\operatornamewithlimits{argmax}}

\subsection{Selecting invariant features}
\label{feature_sel}

The first family of DA methods is based on the selection of invariant features. Invariant features are usually a subset of the original set of features, which are \blu{the most} robust to changes from the source to the target domain.
The main idea of the approach is to select features in order to reduce the difference between ${P^s(X,Y)}$ and ${P^t(X,Y)}$. 
An alternative strategy to encode {\it invariance} is based on the inclusion of additional (synthetic) labeled samples in the training set, a procedure known in machine learning as {\it data augmentation}. A method adopting this strategy was studied in~\cite{Ver13}, where sample selection bias problems are addressed by enriching the training set with artificial examples that correspond to physical consistent variations of the training samples (illumination, size, rotation). To limit the number of additional examples to be used by the SVM, variations are generated only for training samples considered as support vectors by the classifier trained on the source domain only.

Let us consider \blu{the first strategy, and focus on the} analysis of hyperspectral images as an application of particular interest. 
Hyperspectral sensors are capable to capture hundreds of narrow spectral bands from a wide range of the electromagnetic spectrum. For this reason,  
\blu{they are particularly sensitive to subtle changes in the image acquisition conditions, leading to a non-stationary behavior of the spectral signature of the classes and therefore to problems that should be solved by transfer learning and DA approaches.} An example of shift in the signature of an hyperspectral image acquired by the Hyperion sensor over two areas of the Okavango Delta in Botswana is provided in Fig.~\ref{fig:BOT}.

In \cite{Bruzzone2009}, the authors propose an approach for selecting subsets of features that are both 1) discriminative of the land-cover classes and 2) invariant between the source and the target \blu{domain}.
The main idea of this approach is to explicitly consider two distinct terms in the criterion function for evaluating both 1) the {\it discrimination capability} $\Delta$ of the feature subset, and 2) the {\it data-set shift} $P$ of the features between source and target domain. 
The first term is standard in filter methods for feature selection \blu{and provides high scores when the features selected show some kind of dependency with the desired output (e.g. the classes to be predicted)}.
The second term has been introduced to evaluate the {\it invariance} of the feature subset between the two domains.
The subset of features $F$ is selected by jointly optimizing the two terms $\Delta$ and ${P}$, i.e., by solving the following multi-objective optimization problem:
\begin{equation}
\argmin_{|F|=l} {(-\Delta(F), P(F))},
\label{eq:MOOP1}
\end{equation}
where $l$ is the size of the feature subset.
Both $\Delta$ and ${P}$ are treated as functions of the subset of considered features $F$.
 The specific definition for the terms $\Delta$ and ${P}$ are reported in \cite{Bruzzone2009} considering their parametric estimation (assuming Gaussian distribution of the classes) in both the supervised and semisupervised DA setting.
In \cite{Persello2015}, the two terms are defined considering kernel-based dependence estimators and kernel embedding of conditional distributions resulting in a nonparametric approach, which does not require the estimation of the class distributions as an intermediate step. 
Problem~\eqref{eq:MOOP1} is solved by adopting a genetic multi-objective optimization algorithm.
The solution results in features with high capability to discriminate classes (small value of $-\Delta$) and high stability on the two domains (small data set shift ${P}$).
Adopting a multi-objective optimization approach instead of considering a linear combination of the two terms frees the user from specifying in advance the relative importance of the two terms $\Delta$ and ${P}$. 
The solution of the multi-objective problem allows one to find the solutions that represent the best trade-offs of discriminative and stable feature subsets for the specific transfer learning problem at hand.

{\blu{In the following, w}e report the experimental results obtained on a hyperspectral image acquired by the Hyperion sensor of the Earth Observation 1 satellite in an area of the Okavango Delta, Botswana \cite{Ham2005} (see Fig.~\ref{fig:BOT}). 
For more information about the experimental setting and the obtained results we refer the reader to \cite{Bruzzone2009}.
The labeled reference samples were collected on two spatially disjoint areas with slightly different characteristics, thus representing two different domains. The samples taken on the first area (considered as source domain) were partitioned into a training set $T^s$ and a test set $\mathcal{T}^s$ by a random sampling. Samples taken on the second area (target domain) were used to derive a training set $T^t$ and test set $\mathcal{T}^t$  according to the same procedure. 
\begin{figure}[t]
\centerline{\includegraphics[width=0.85\columnwidth]{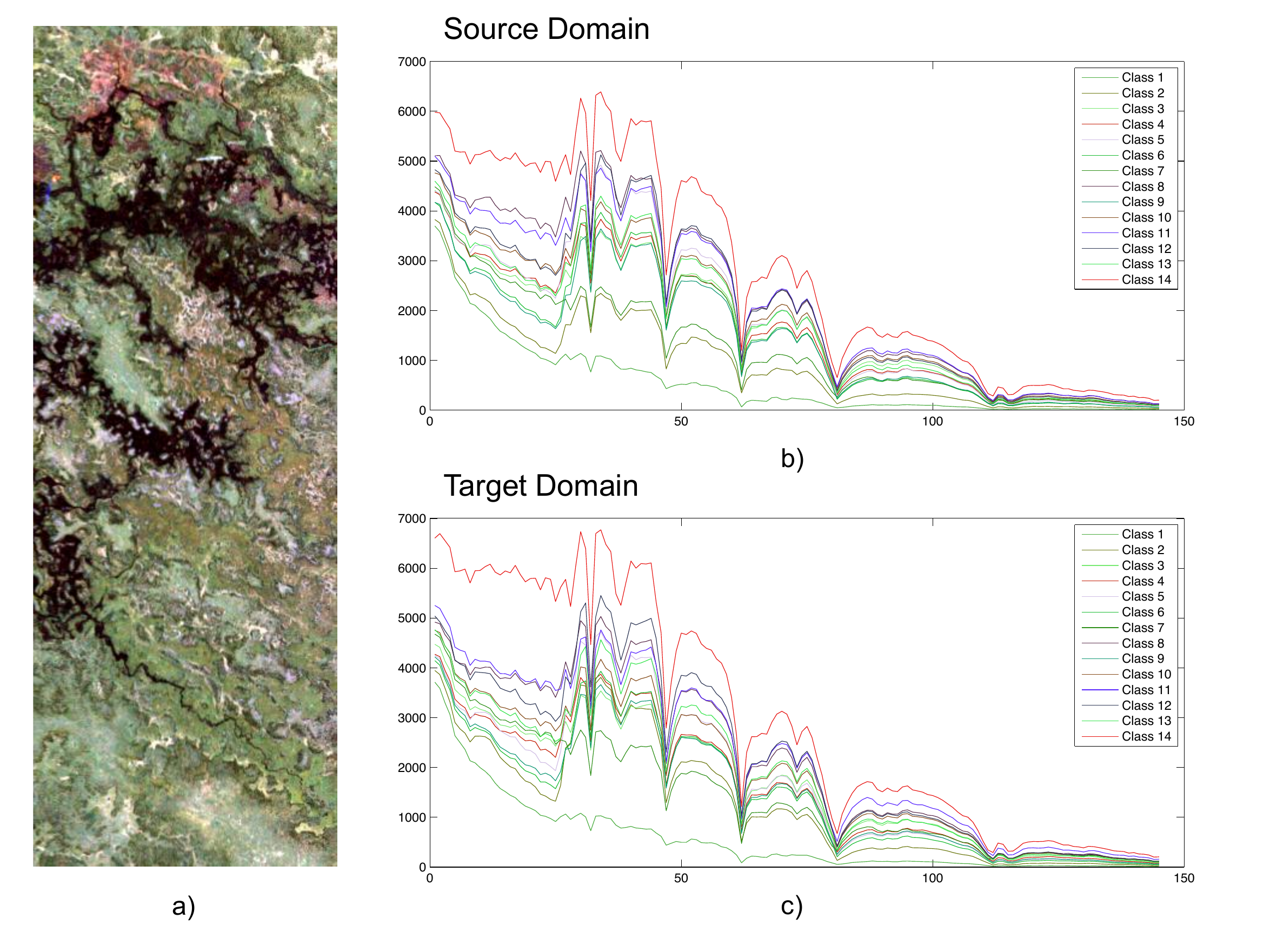}}
\caption{a) False color composition of a portion of the hyperspectral data set. b) Mean spectral signature of the classes on the source domain. c) Mean spectral signature of the classes on the target domain.}
 \label{fig:BOT}
 \end{figure}
The estimated Pareto front for the selection of $6$ features is reported in Fig.~\ref{fig:graphs}. \blu{Each point in the graphs  corresponds to a different feature subset $F$ selected (i.e., a feature set minimizing Eq.~\eqref{eq:MOOP1}).} In panel a), the color of the points indicates the Overall Accuracy (OA) obtained on the source-domain test set $\mathcal{T}^s$ using an SVM classifier trained using $T^s$ (according to the reported color scale bar). In panel b), the color indicates the OA obtained by the \blu{same} SVM classifier on the target-domain test set $\mathcal{T}^t$. 
\begin{figure}[tb]
\begin{minipage}[b]{.9\linewidth}
	\centering
	\centerline{\includegraphics[width=1\linewidth]{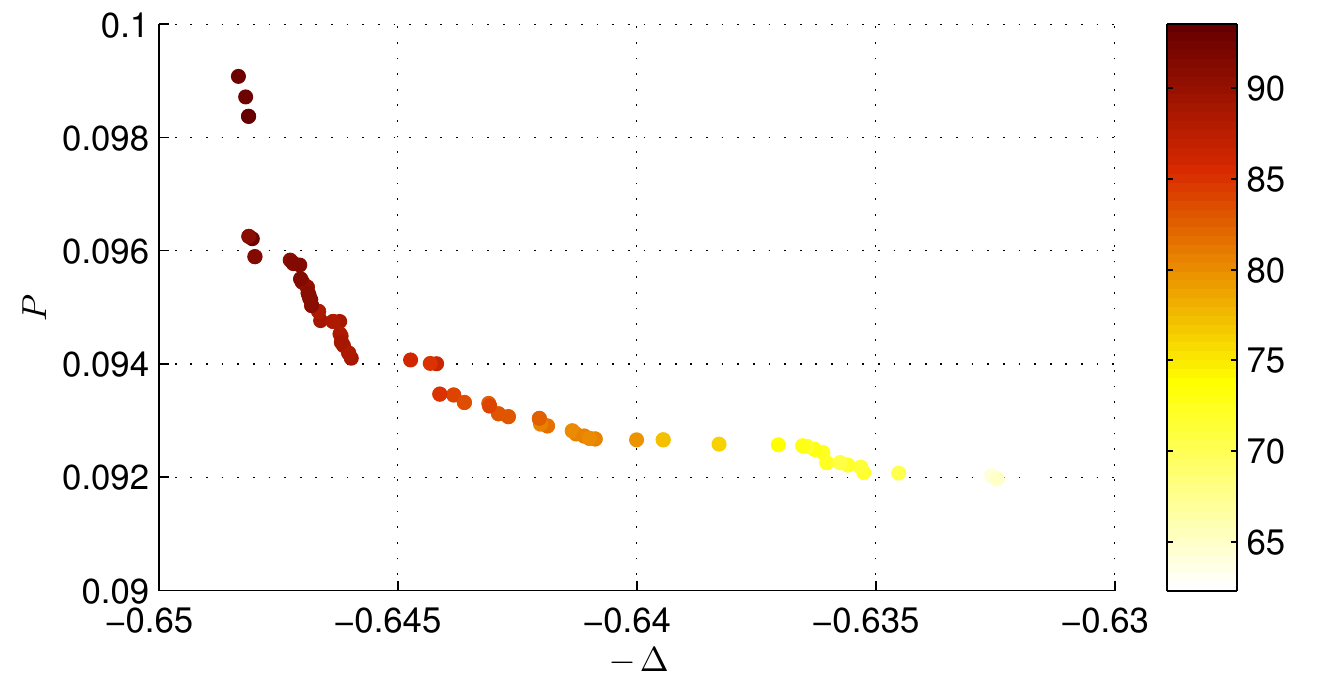}}

  \centerline{a)}\medskip
\end{minipage}

\begin{minipage}[b]{0.9\linewidth}
  \centering
	\centerline{\includegraphics[width=1\linewidth]{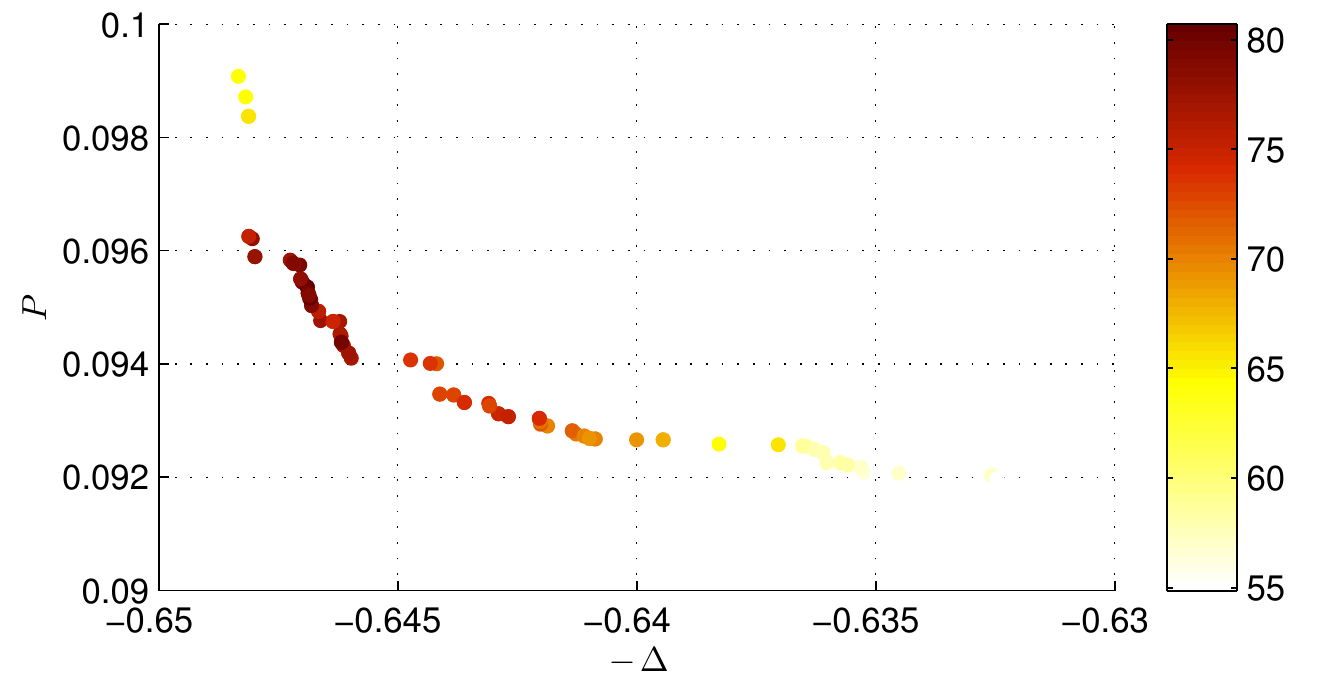}}

  \centerline{b)}\medskip
\end{minipage}
\caption{Pareto front estimated using a multi-objective genetic algorithm for the selection of ten features. \blue{Each dot corresponds to a feature set minimizing Eq.~\eqref{eq:MOOP1}}. The color indicates the overall accuracy on a) source test set $\mathcal{T}^s$ and b) target test set $\mathcal{T}^t$ according to the reported color scale bar \blue{(adapted from \cite{Bruzzone2009})}.}
 \label{fig:graphs}
 \end{figure}
The results show that the solutions with higher relevance $\Delta$ result in better classification accuracies on the source domain. 
However,  relevance only is not sufficient for selecting features that are stable for the classification on the target domain.    
We observe that the most accurate solutions on the target domain $\mathcal{T}^t$ are those that exhibit a good tradeoff between the relevance and invariance terms. 
This confirms the importance of the invariance term and shows that the $P$ measure is able to capture the information of feature stability.
In order to select the subset of features that leads to good generalization capabilities on different domains, tradeoff solutions between the two competing objectives should be identified.
The selected subset of features results in an OA of 91.0\% on the source domain and 80.7\% on the target.
The set of features selected according to the optimization of $\Delta$  resulted in an OA of 92.7\% on the source\blu{, but only of} 64.4\% on the target.
This result shows that the features selected by accounting for the dataset shift between the domains can significantly improve the generalization capability on the target domain.

\subsection{Adapting data distributions}\label{sec:represent}

\blu{The second family reviewed considers} DA methods aiming at adapting the representation of the original data, regardless of the model that will process them afterwards.
A review of the methods proposed in computer vision and machine learning can be found in~\cite{Pat15}. Here we will focus on the approaches proposed in remote sensing literature. This type of adaptation is often done by relative normalization methods, i.e., methods that do not provide physical units as an output, rather similarly distributed digital numbers. Their aim is to make the data distributions more similar across the domains in order to train a single model that can classify simultaneously the source and target domain(s). 

In general, a data representation transformation with the aim of making data sources more similar should have the following desirable properties.

\begin{itemize}
\item[-] The method should be able to align unpaired data ({\bf Unpaired}): this allows to align non coregistered data (not even imaging the same location), or data with different spatial resolutions.
\item[-] The method should be able to align data of different dimensionality ({\bf $\Delta$ dim.})  to allow multisource classification.
\item[-] The method should be able to align several domains at the same time ({\bf Multisource}), to enhance multitemporal adaptation, instead of pairwise adaptation.
\item[-] The method should be able to align in a nonlinear way ({\bf Nonlinear}), since the transformation between domains can be nonlinear because of atmospheric or illumination effects.
\item[-] The method should be able to use labeled information from the source domain, when available ({\bf Labels in $s$}). A discriminative transform tends to align better the datasets, since it aligns the data according to the semantic \blu{classes} required by the user.
\item[-] The method should avoid to be forced to use labeled information in all domains ({\bf No labels in $t$}), as labels might not be available in all domains or their acquisition would have a high cost (typically through terrestrial campaigns, see Section~\ref{sec:AL}).
\end{itemize}

Several methods have been proposed in recent \blu{remote sensing} literature. We provide a brief review below and a summary in Table~\ref{tab:summary}. Depending on the specific situation, the analyst can use this table to select the most suitable approach.

Most of the recent literature focuses on feature extraction strategies, where the extracted features align the data spaces to each other. In that space, the same classifier (or regressor) can be applied to  all the domains. Beyond works dealing with traditional or multidimensional histogram matching~\cite{Ina08} or data alignment with PCA or kPCA~\cite{Nie09}, in~\cite{Mat13c} authors propose to \blu{minimize the statistical distance between domains, which is assessed through  a kernel-based dependence estimator, the Maximum Mean Discrepancy (MMD~\cite{Mat14b})}. Other studies still focus on feature extraction, but based on multiview models: in~\cite{Nie07}, Nielsen aligns domains with canonical correlation analysis (CCA) and performs change detection therein. The approach is extended to a kernel and semisupervised version in~\cite{Vol14b}, where the authors perform  change detection with different sensors. In~\cite{Tui13d}, the domains are matched in a latent space defined through an eigenproblem aiming at preserving label (dis)similarities and the geometric structure of the single manifolds. \blue{A nonlinear (kernelized) version of the algorithm has been also proposed in~\cite{Tui15d}}. This approach is particularly appealing, since it can align an arbitrary number of domains of different dimensionality (as (k)CCA), but without requiring paired examples. However, it has the disadvantage of requiring labeled samples in all domains.  In \cite{Mar15}, the authors relax this requirement by working on semantic ties, i.e. samples issued from the same object, but whose class is unknown. This last method therefore requires at least a partial overlap between the images in order to find the ties, either manually or by stereo matching, as in \cite{Mon13}. \blue{Authors in~\cite{Yan16} regularize the manifold alignment solution with spatial information, leading to a more stable feature representation transfer. In~\cite{Yan16b} they propose a multiscale approach considering the preservation of both local and global geometric characteristics and relying on clustering pairs, rather than labeled correspondences}. \blue{Other recent methods rely on eigendecompositions, such as those proposed in~\cite{Sun15,Sun16}. In~\cite{Sun15} two PCA eigensystems (one for the source and another for the target domain) are aligned by minimizing their divergence. In~\cite{Sun16}, authors consider a sparse representation approach, where they reduce the difference between domains again by minimizing the MMD. In both papers~\cite{Sun15,Sun16}, the authors aim at transferring categories models learned on landscape views to aerial views from VHR remote sensing images.}
\blu{In \cite{Zhang2014}, the authors propose a set of techniques based on sample reweighing and transformation to address different DA situations. The study offers also a causal interpretation of the different forms of domain shift. The adaptation strategies are developed on the basis of the embedding of sample distributions in the reproducing kernel Hilbert space.}
\\
Beyond classical feature extraction, authors in \cite{Pet11} align multitemporal sequences based on a measure of similarity between sequences barycenters, which corresponds to a global alignment of the spectra in a time series of images. In \cite{Jun11}, authors consider spatial shifts in large image acquisitions: the spectra are spatially detrended using Gaussian processes  to avoid shifts related to localized class variability. \blue{In \cite{Zha14}, authors perform anomaly detection by a sparse discriminative transform that i)  maximizes the distance between the anomaly class and the background classes (defined as a set of endmembers) and ii) minimizes the distance between the source and target distributions after reduction by PCA.} In \cite{Tui12a} the authors consider the domains as multidimensional graphs and propose to align the domains by solving a graph matching problem. \blu{Finally, authors in~\cite{Oth16} find a multispectral mapping between source and target spectra, in order to project the labeled pixels of the source into the target domain: tie points are found between the labeled  source pixels and the pixels in the target by registration and then the mapping between source and target is learned by regression between the corresponding pairs. Then, the labeled pixels are projected into the target domain and are used to train a classifier therein. As for~\cite{Mar15}, partial overlap between the images is required.}

As one can see in Table~\ref{tab:summary}, some methods will be more suitable than others depending on the problem: for example canonical correlation-based methods can be used only for coregistered data, while non-multiview methods ((k)PCA, (SS)TCA) cannot align more than two domains at a time.

\begin{table*}[t!]
\begin{center}
\caption{Representation alignment methods used in remote sensing.} 
\label{tab:summary}

\setlength{\tabcolsep}{5pt}
\begin{tabular}{|l|c|c|c||cc|c|}
\hline
\hline
{\bf Method}                   & {\bf Labels in $s$} & {\bf No labels in $t$} & {\bf Multisource} & {\bf Unpaired} & {\bf $\Delta$ dim.} & {\bf Nonlinear}  \\
\hline\hline
PCA       & $\times$ & $\checkmark$& $\times$ & $\checkmark$  & $\times$ & $\times$  \\\hline
kPCA~\cite{Nie09}      & $\times$ & $\checkmark$& $\times$ & $\checkmark$ & $\times$  & $\checkmark$  \\\hline
(SS)TCA~\cite{Mat13c}             & $\times \checkmark$ & $\checkmark$& $\times$ & $\checkmark$  & $\times$& $\checkmark$   \\\hline
\hline
CCA~\cite{Nie07}     & $\times \checkmark$ & $\times \checkmark$& $\checkmark$  & $\times$ & $\checkmark$ & $\times$  \\\hline
kCCA~\cite{Vol14b}            & $\times \checkmark$ & $\times \checkmark$& $\checkmark$  & $\times$ &  $\checkmark$ & $\checkmark$   \\\hline
MA~\cite{Yan16b}              & $\checkmark$ & $\checkmark$& $\checkmark$  & $\checkmark$  &$\checkmark$ & $\times$  \\\hline
SSMA~\cite{Tui13d}           & $\checkmark$  & $\times$  & $\checkmark$ & $\checkmark$  &$\checkmark$ & $\times$  \\\hline
KEMA~\cite{Tui15d}           & $\checkmark$  & $\times$  & $\checkmark$ & $\checkmark$  &$\checkmark$ & $\checkmark$  \\\hline
\hline
GM~\cite{Tui12a}             & $\times$ & $\checkmark$ &$\times$& $\checkmark$  & $\times$ & $\times$  \\\hline
\end{tabular}\\
\end{center}
\end{table*}

In the following we compare a series of methods on the challenging problem of transferring a classifier over a multiangular sequence of images over Rio de Janeiro~\cite{Pac11} illustrated in Fig.~\ref{fig:ang}. More details on this example can be found in~\cite{Tui13d}. The images are not coregistered, but are all acquired over a single pass of the WorldView2 sensor. For this reason, the only shifts observed are due to angular effects. The problem is a 11-classes problem, and a separate ground truth is provided per each image (Table~\ref{tab:ang}).

\begin{table}[!t]
\caption{Number of labeled pixels available for each dataset in the multiangular experiments ($\theta$ = off-nadir angle).}
\label{tab:ang}
\setlength{\tabcolsep}{4pt}
{\small
\begin{tabular}{c|c|c|c|c|c}
\hline
\backslashbox{Class}{$\theta$} & $-38.79^\circ$ & $ -29.16^\circ$ & $ 6.09^\circ$ & $ 26.76^\circ$ & $ 39.5^\circ$ \\\hline\hline
Water & 83260 &79937& 66084& 63492 & 54769 \\
Grass & 8127 &8127& 8127& 8127 & 8127 \\
Pools & 244&244& 223 & 195 & 195 \\
Trees & 4231 &4074& 3066& 3046 & 3046 \\
Concrete & 707 &719&719 & 719 & 696 \\
Bare soil & 790 &790& 790& 790 & 811 \\
Asphalt & 2949 &2949& 2949& 2827 & 2827 \\
Grey buildings & 6291 &6061& 5936& 4375 & 4527\\
Red buildings & 1147&1080& 1070& 1046 & 1042 \\
White buildings & 1683 &1683& 1571& 1571 & 1571 \\
Shadows & 1829 &1056& 705& 512 & 525 \\
Tarmac & 5179 &5179& 5179& 2166 & 2758 \\
\hline
\end{tabular}
}
\end{table}

\begin{figure*}[!t]
\centering
\begin{tabular}{ccccc}
$- 38.79^\circ$ &  $-29.16^\circ$ & $6.09^\circ$ & $26.76^\circ$ & $39.5^\circ$ \\
\includegraphics[width=3cm]{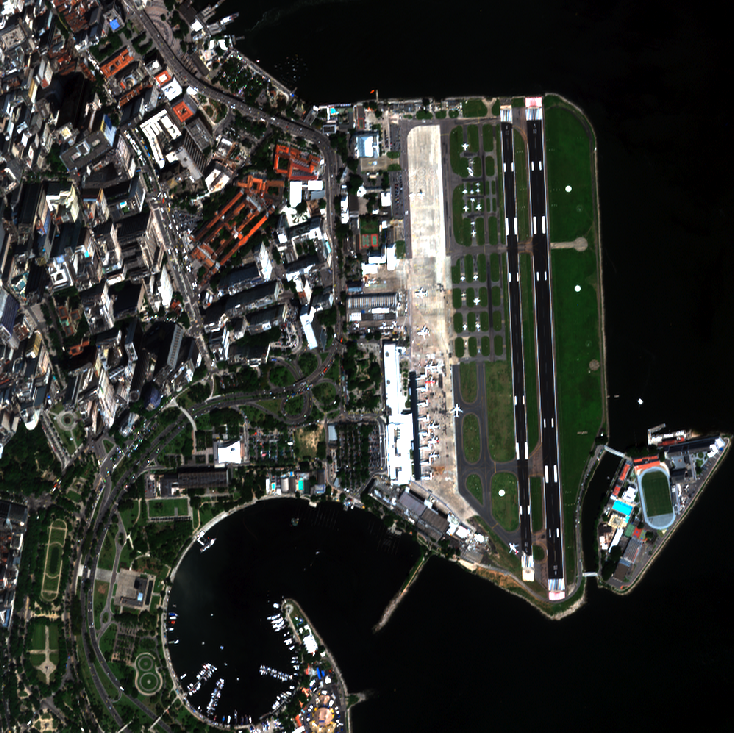}&
\includegraphics[width=3cm]{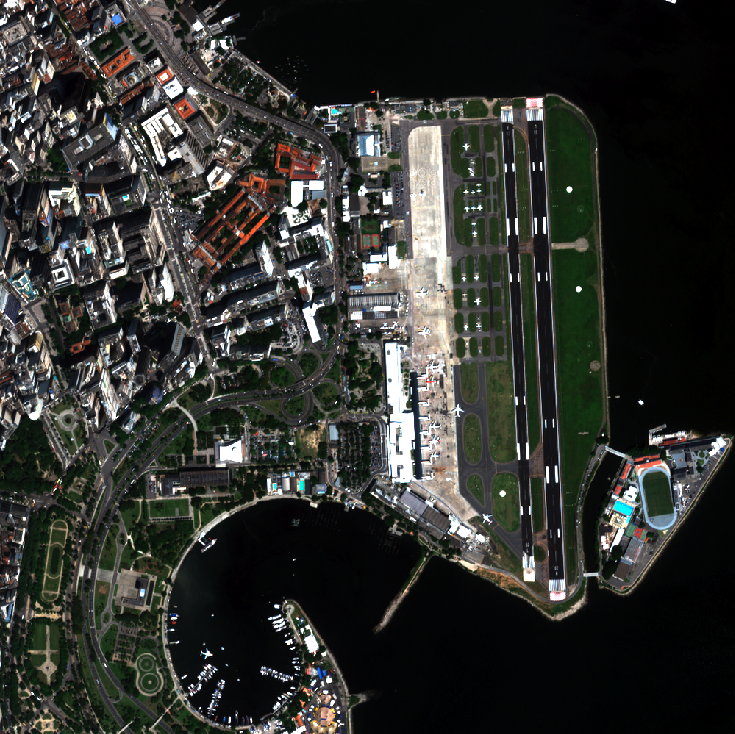}&
\includegraphics[width=3cm]{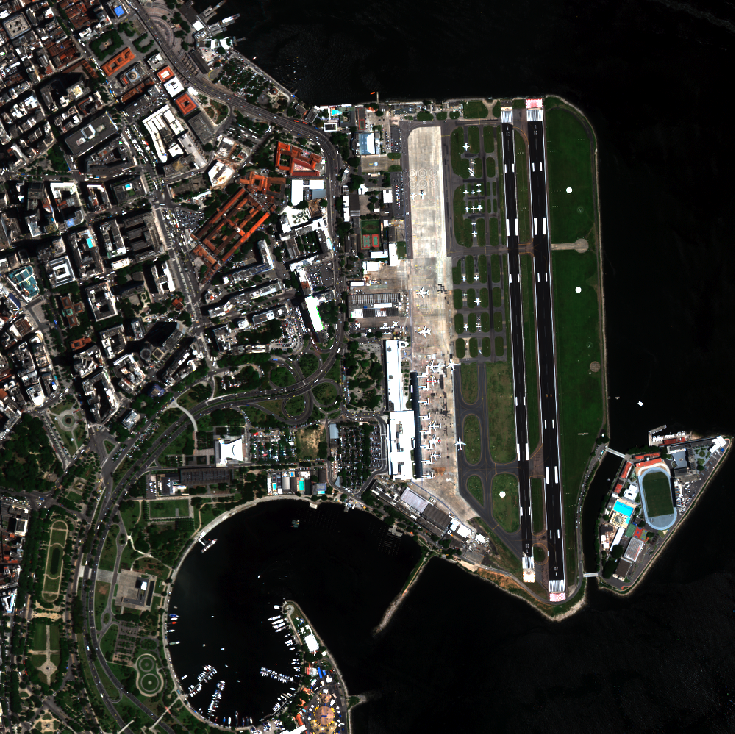}&
\includegraphics[width=3cm]{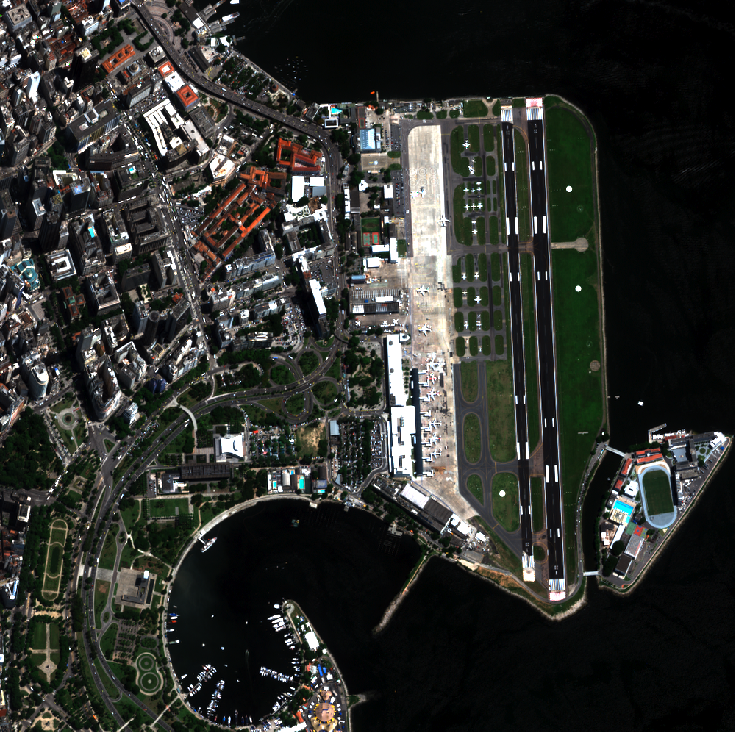}&
\includegraphics[width=3cm]{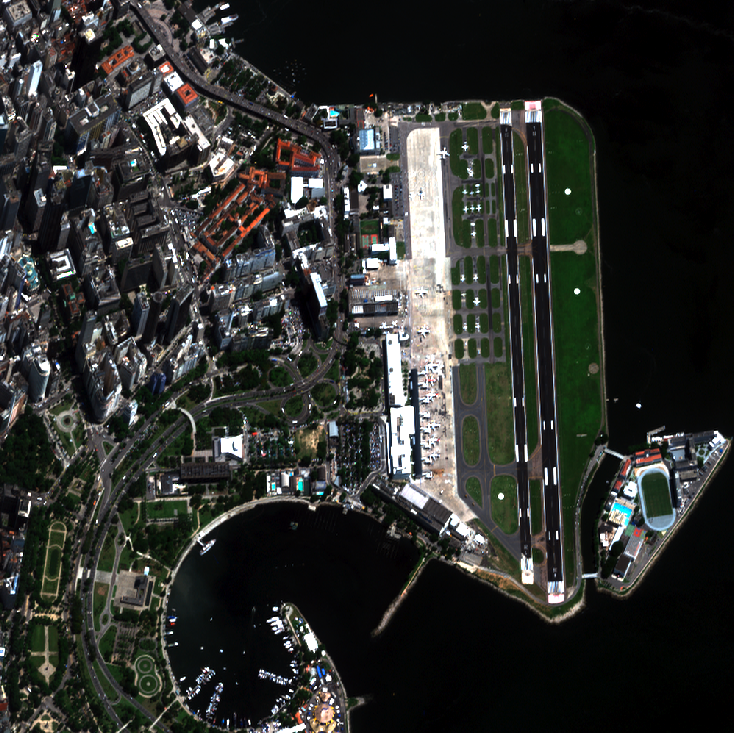}\\
\end{tabular}
\caption{The five images of the Rio de Janeiro angular sequence~\cite{Pac11}.}
\label{fig:ang}
\end{figure*}

The adaptation experiment is designed as follows: we take the nadir image (off-nadir angle $\theta = 6.09^\circ$) as the source one and use all the other as target ones. We apply the PCA, KPCA, GM and SSMA  transforms  and then train a classifier using 100 labeled pixels from the source domain  and predict all the target domains using that classifier, without further modifications. For PCA, KPCA and GM, the adaptation is done for each target domain separately, while for SSMA a single adaptation projection is obtained for all domains at once. For SSMA we also used 50 labeled pixels per class from each target domain. To be fair in the evaluation,  the projections for the PCA, KPCA and GM methods are obtained in an unsupervised way, but then the classifier is trained using the original training points from the nadir acquisition, stacked to the transformed labeled pixels of the domain to be tested. We also add a best case, where we directly use labeled samples from the target domains for the classification.

The results are illustrated in Fig.~\ref{fig:angres}: predicting in the off-nadir images using the original training samples from the nadir image leads to poor results, especially for strong off-nadir angles. All the methods considered leverage the decrease in performance and lead to a quasi-flat prediction surface (meaning that the model can predict correctly, regardless of the angular configuration), with particular good performances for SSMA method, which seems to align at best the data distributions. This is not surprising, since among the  tested methods SSMA is the only one with a clear discriminative component (it uses labels in all domains to define the projections).

\begin{figure}[!t]
\centering
\begin{tabular}{cc}
\includegraphics[width=.45\linewidth]{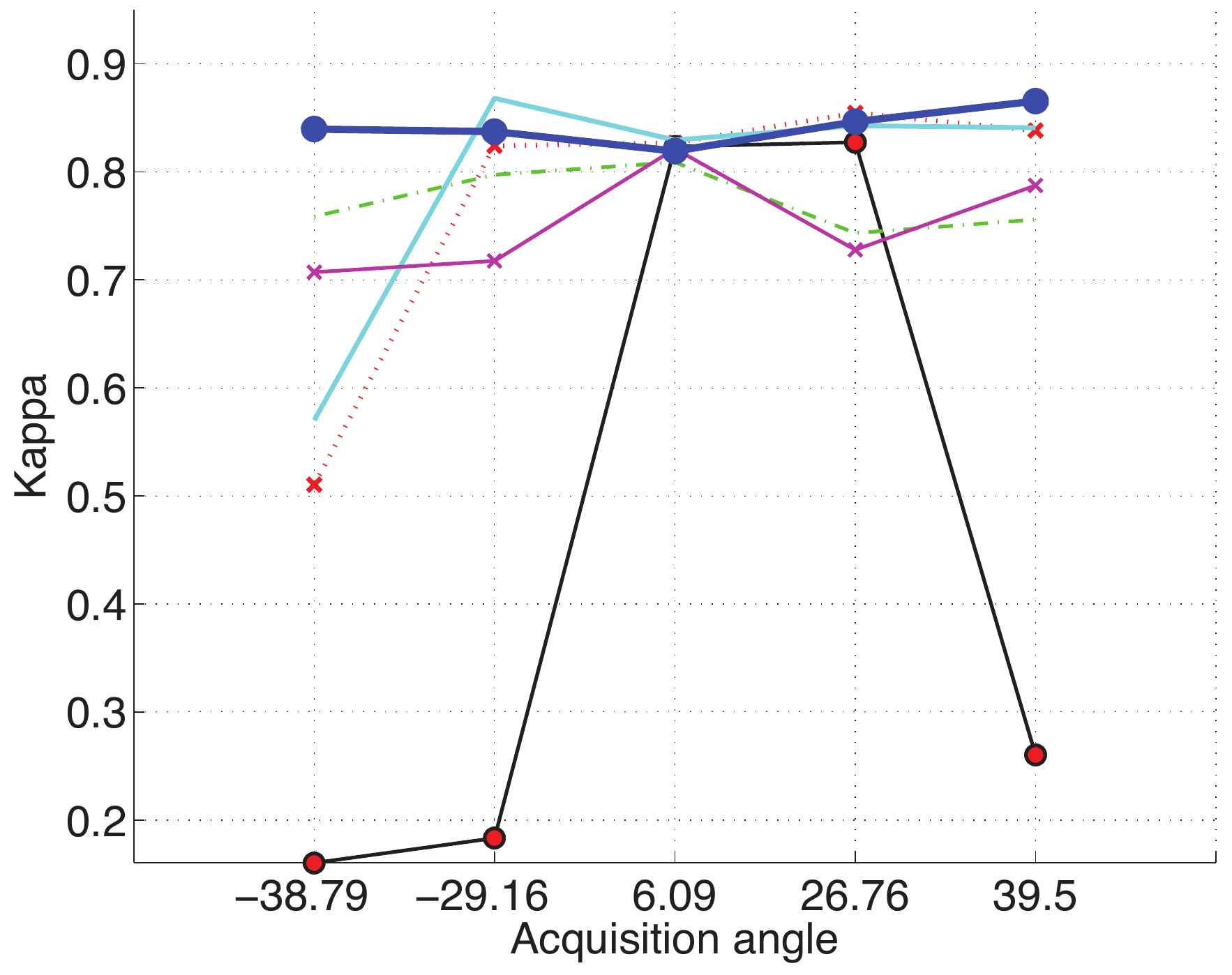}&
\includegraphics[width=.45\linewidth]{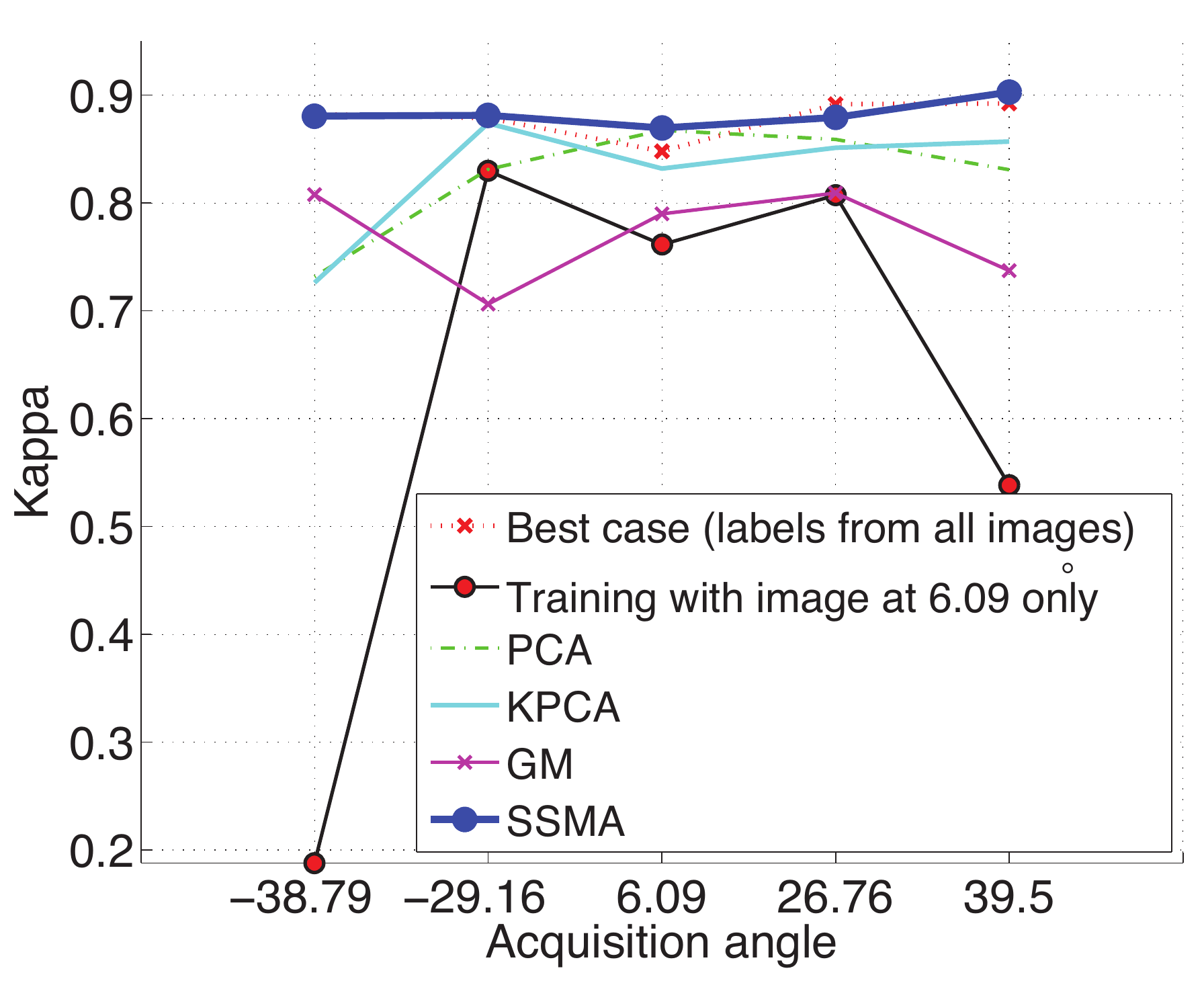}\\
LDA & SVM \\
\end{tabular}
\caption{Classification results over the five Rio acquisitions (adapted from~\cite{Tui13d}). 100 labeled samples per class from the nadir image are used to train a classifier then used to test on the others. In the SSMA experiment, 50 labeled pixels per class are used from the other acquisitions.}
\label{fig:angres}
\end{figure}

\subsection{Adapting classifiers with semisupervised approaches}

A widely used approach to DA is based on the adaptation of the model of the classifier derived on the source domain to the target domain. In the literature the approach is defined semisupervised if this adaptation is based only on unlabeled samples of the target domain, i.e. no target training samples are used. The rationale of semisupervised adaptation is to use the relations between the distributions of the source and target domains in order to infer a reliable solution to the problem described in the target domain. The common assumption of most of the methods is that the source and target domains share the same set of classes \blu{and features}.

The first attempts to address semisupervised domain adaptation in remote sensing image classification have been presented in \cite{Bru01}, where a domain adaptation technique is proposed that updates the parameters of an already trained parametric maximum-likelihood (ML) classifier on the basis of the distribution of a new image for which no labeled samples are available. In~\cite{Bru02}, the ML-based domain adaptation technique is extended to the framework of the Bayesian rule for cascade classification (i.e., the classification process is performed by jointly considering information contained in the source and the target domains). The basic idea in both the methods is modeling the observed spaces by a mixture of distributions, whose components can be estimated by the use of unlabeled target data. 
This is achieved by using the EM algorithm with finite Gaussian Mixture Model. In~\cite{Bru02b,Bru02c}, domain adaptation approaches based on multiple-classifier and multiple-cascade-classifier architectures have been defined. \blu{As base classifiers} Gaussian ML classifiers, radial basis function neural-networks and hybrid cascade classifiers \blu{are used}. Decision trees update and randomization are used for domain adaptation in~\cite{Raj06b}: a set of decision trees is made robust to dataset shift either by training it with the EM using density functions from the target domain (similarly to~\cite{Bru02}) or by randomizing the decision trees (but in this case without control on the  adaptation objective); authors also propose a semisupervised extension, where the classifiers performing poorly on the (few) labeled samples in the target domain are downweighted in the final decision. Finally, the domain adaptation technique proposed in~\cite{Bah12} for multitemporal images addresses the challenging situation where source and target domains have a different set of classes. The sets of classes of the target and the source domains are automatically identified in the DA step by the joint use of unsupervised change detection and Jeffreys-Matusita statistical distance measure. This process results in the detection of classes that appeared or disappeared between the domains. 

The semisupervised problem has also been extensively studied in the framework of kernel methods with Support Vector Machine (SVM) classifiers. This has been done especially for  addressing sample selection bias problems \blu{(see Section~\ref{sec:DA})}. 
Most of the semisupervised techniques proposed with SVM exploit the cluster assumption, i.e., adapt the position of the hyperplane estimated on the source domain to the target domain assuming that it should be located in low density regions of the feature space. In~\cite{Bru06}, \blu{authors employ the Transductive SVM, a method that iteratively moves the decision boundary of the SVM classifier towards low-density areas of the (unlabeled) target domain.}
Other semi-supervised approaches are imported in remote sensing in~\cite{Chi07}, where the SVM semisupervised learning is addressed in the primal formulation of the cost function. \blue{In~\cite{Sun13}, authors regularize the support vector machine solution by adding a new term in the optimization accounting for the divergence between source and target domain (the MMD discussed in Section~\ref{sec:represent}). By doing so, the decision function selected depends on a kernel that both projects in a discriminative space and minimizes the shift between training and test data. In~\cite{Leiva-Murillo2013}, authors cast the domain adaptation problem as a multi-task learning problem, where each source-domain pair (each task) is solved by deforming the kernel by sharing information among the tasks.} The Laplacian SVM technique applied to the classification of multispectral remote sensing images is presented~\cite{Gom08}. It exploits an additional regularization term on the geometry of both the labeled and the unlabeled samples by using the graph Laplacian. In~\cite{Kim10}, authors also use a manifold-regularized classifier in a semisupervised setting, where the adaptation is performed by adding semilabeled examples from the target domain.

A specific semisupervised SVM defined for addressing DA problems is presented in~\cite{Bruzzone2010}. 
The \blu{Domain Adaptation SVM (}DASVM\blu{)} starts from a standard supervised learning on the training samples of the source domain that is followed by an iterative procedure. At each iteration it includes in the learning cost function a subset of unlabeled samples of the target domain adequately selected, while gradually removes the training samples of the source domain. At convergence the DASVM can classify accurately the samples of the target domain.

\subsection{Adaptation of the classifier by active learning}
\label{sec:AL}

In most of the above-mentioned approaches, it is assumed that no label information can be obtained in the newly acquired (target) domains (semisupervised DA). This assumption may hinder the success of classification in the case of very strong deformations or when new classes, unseen during training, appear in the test data. A small amount of labeled data issued form the target domain may solve this problem efficiently. 
However, since the acquisition is timely and can be costly, it becomes mandatory to chose the samples well. Active learning strategies have been proposed to tackle this challenging tasks and guide the DA process with the selection of the most informative target samples \cite{Jun2008,  Raj08b, Per12, Persello2013, Tui11d, Mat12}. 

Active learning is the name of a set of methodologies aiming at the interaction between a user and a prediction model, where the first provides labels by his / her knowledge of the \blu{task to be solved} and the second performs the \blu{prediction} and highlights samples, for which it has the highest uncertainty~\cite{Set12}. By focusing on these samples, the user provides the labels where they help the most and thus allows the classifier to migrate in a fast way towards the optimal model. Surveys on active learning methods applied to remote sensing can be found in~\cite{Tui11,Cra12, Bruzzone2012}. In \blu{the case of domain adaptation}, the user provides examples coming from the target domain only and the optimal classifier is the one that would have been obtained with several examples in the target domain. The process starts for a classifier that is optimal for the source domain, and gradually evolves to model the data distribution in the target domain. Figure~\ref{fig:AL} summarizes the AL process for domain adaptation.

\begin{figure}
\includegraphics[width= \linewidth]{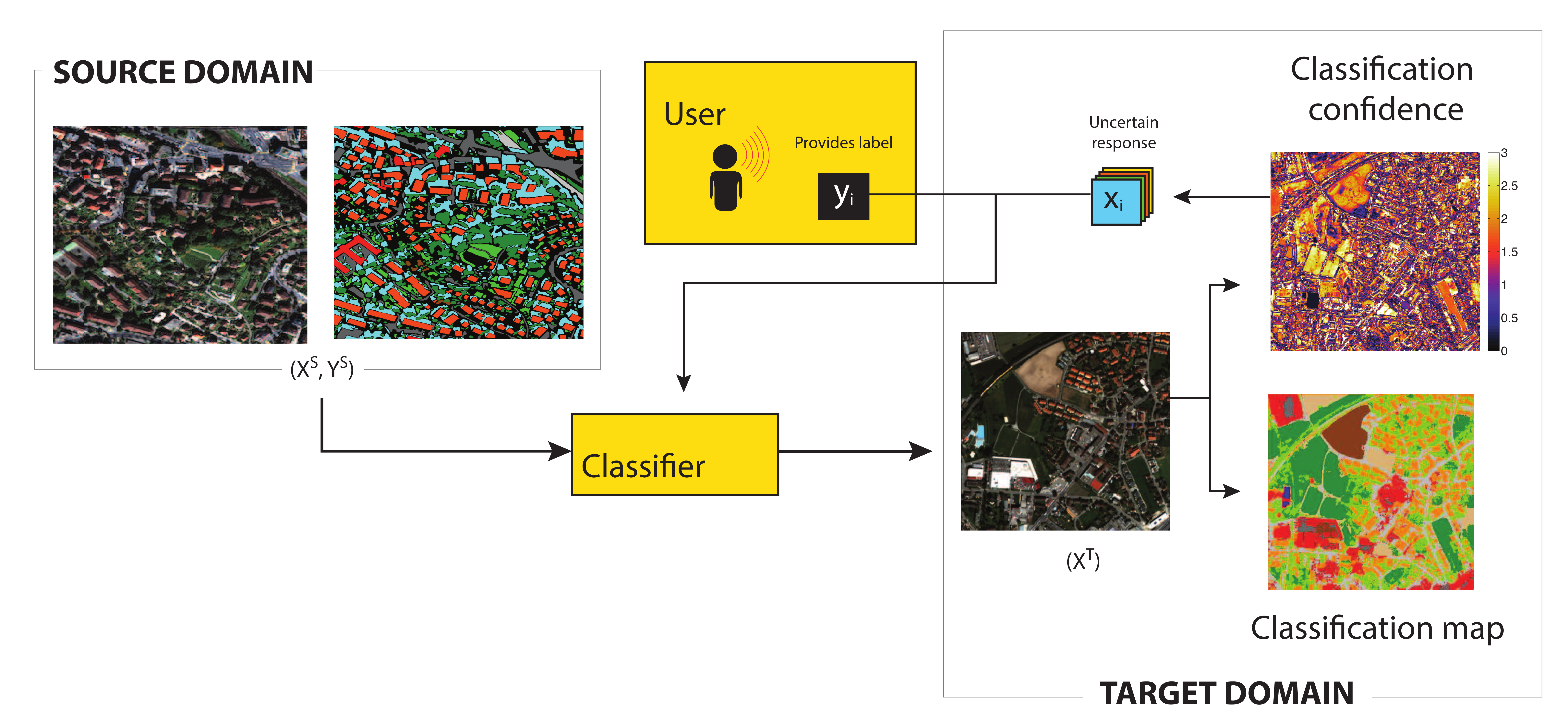}
\caption{Flowchart of the active learning paradigm for domain adaptation.}
\label{fig:AL}
\end{figure}

 One could apply classical active learning strategies in transfer learning problems (under the sample selection bias assumption) with successful results, since classical AL will point out samples close to the current decision boundaries and the user, by the labels provided, will disclose the shifted areas, where the next iterations will focus. \blu{However}, depending on the degree of transformation between the domains, one can use  more sophisticated strategies that take into account measures of the deformation between the domains. In this respect, the following problems have driven DA-related research in active learning:

\begin{itemize}
\item When it is expected that new classes appear in the target domain, active learning can be used to highlight the areas of the feature space where these classes could be. By using the reasoning of sample selection bias, in~\cite{Tui11d}, the feature space in the target domain is screened using clustering, and dense clusters with no labeled samples are presented to the user, who can then provide labels if new classes are present. In~\cite{Dem12}, the detection of new classes is set as a change detection problem, where uncertainty of changes is assessed with an information theoretic criterion. Image time series are analyzed in~\cite{Dem13}.
\item  When significant differences between source and target domains are expected (i.e., the sample selection bias assumption does not hold), the presence of labeled source samples (although beneficial at the beginning of the process), can be harmful for the classification of the target domain \cite{Per12, Persello2013}. Refer to the example discussed in Section~\ref{sec:DA}, page~\pageref{sec:classOverlap}: if the distributions of the classes in the target domain overlaps with those of different classes in the source domain,  relying on the labels from the source will lead to errors of the classifier. Accordingly, approaches in \cite{Per12, Mat12,Persello2013} consider reweighting of the samples in the training set enriched by AL: when samples from the source domain become less relevant or misleading for the correct classification of the target domain, they are downweighted in the adapted classifier or completely removed.  Accordingly, the classifier specializes to the target domain through the inclusion of target samples and gradually {\it forgets} the initial source domain. 
\item When the areas to be processed become very large, specific solutions must be designed to avoid too many iterations of the AL process. In this respect, solutions based on selection of clusters~\cite{Stu12}, \blue{compressed sensing~\cite{Roy15}} or on geographically distributed search strategies~ \blu{\cite{Ala14,Bla14,Dem14,Persello2014c} have  been} considered.
\end{itemize}

\begin{figure}
\begin{tabular}{cc}
\includegraphics[width=.4\linewidth]{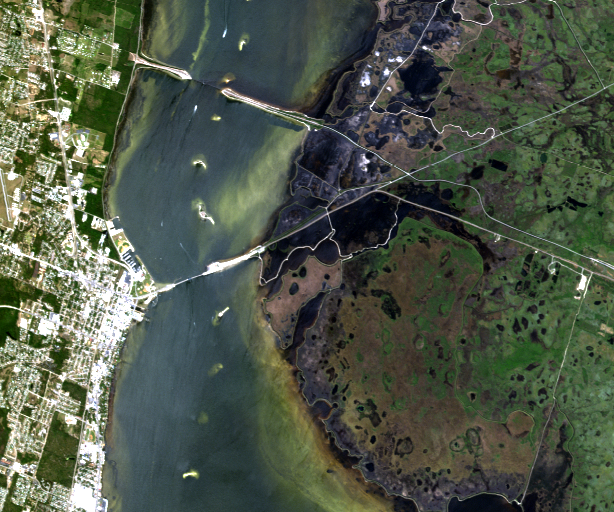} &
\includegraphics[width=.4\linewidth]{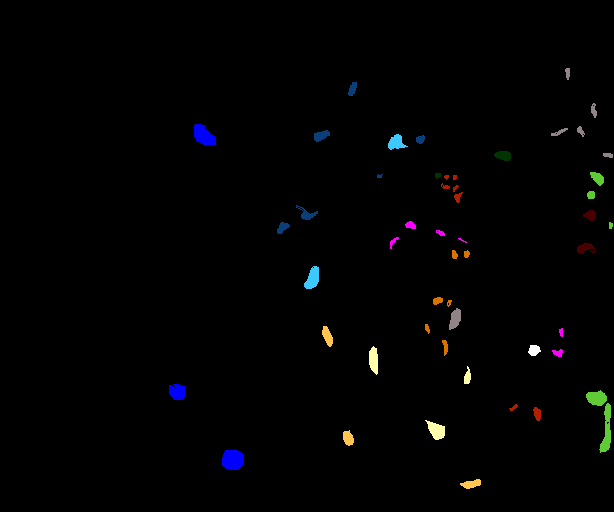} \\
Source image & Source GT\\
\includegraphics[width=.4\linewidth]{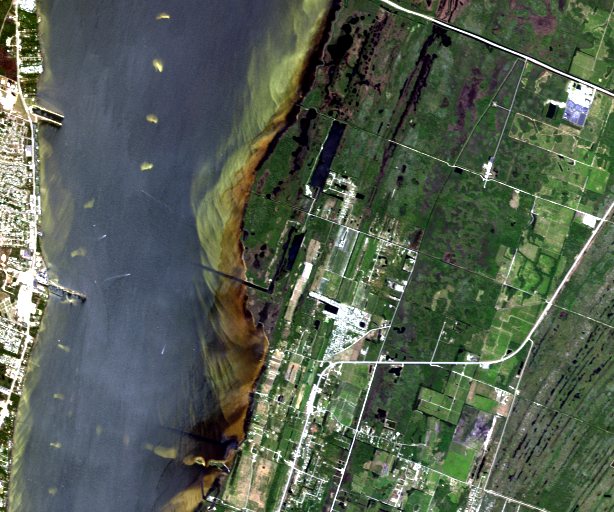} &
\includegraphics[width=.4\linewidth]{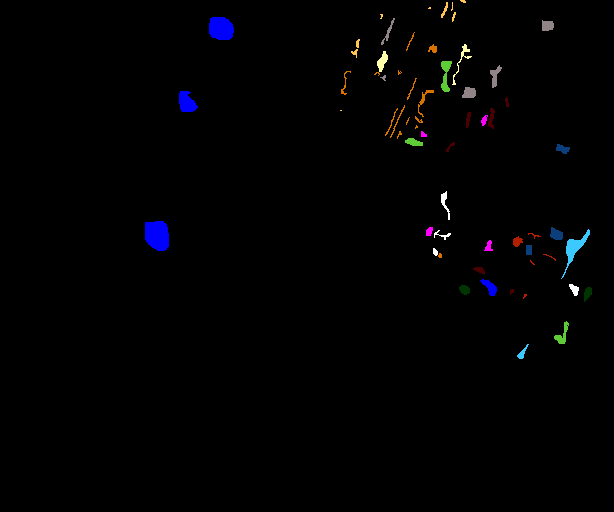} \\
Target image & Target GT (NOT AVAILABLE) \\
\end{tabular}
\caption{Kennedy Space Center data used in the AL domain adaptation experiment.}
\label{fig:KSC}
\end{figure}

In the following, we focus on one example related to the second point above (the reweighting of source samples). This example is adapted from~\cite{Mat12}. We study the feasibility of the migration of a model optimized for landcover mapping in a geographical area to another spatially disjoint region. To do so, we consider the well-known Kennedy Space Center (KSC) hyperspectral image acquired by the AVIRIS spectrometer (top line of Fig.~\ref{fig:KSC}) and try to adapt the model learned therein to be accurate in a spatially disjoint section of the same flightline (bottom line of Fig.~\ref{fig:KSC}).\\
We consider only the 10 classes present in both images. The starting model is learnt using a training set composed of 50 labeled pixels per class and is then enriched by new samples either added randomly or using the Breaking Ties active learning strategy~\cite{Luo05}. The classifier is an SVM, either standard (when no other mention is done) or adaptive using the TrAdaBoost model, a domain adaptation method based on the reweighting of the SVM sample weights after the inclusion of the new labeled points from the target domain~\cite{Raj08b}. 

When using the source SVM without adaptation, we reach an overall accuracy lower than 65\%, while the results obtained by a SVM trained directly on the target labeled samples (which are available for testing) would provide an  accuracy of 90\% (Fig.~\ref{fig:ALDAres}). Here the shift is clearly visible and relates to a loss in \blu{accuracy} of 25\%. Using a random sampling in the target domain, we get a constant increase in performance (green line with $\textasteriskcentered$  markers), but after 300 queries, we are still 5\% away from the classifier learnt using only 500 samples from the target domain. Moreover, the learning rate is slow and the gain is almost linear with the number of queries. We then assess different domain adaptation strategies. First  TrAdaBoost is applied to the set enriched by the random samples (brown line with $\times$ markers): by forgetting the source domain (i.e. by downweighting the source samples that are contradictory with respect to the new samples from the target domain), we already see a significant improvement that fills half of the gap between the best case and the random sampling. But when using AL (blue line with diamond markers) and even more when using it in conjunction with the TrAdaBoost model (black line with circle markers), the learning rate is much higher in the first iterations (meaning that the first queries are much more effective than in the random sampling experiments) and the model converges to the result obtained with 500 random target queries (\blu{solid} blue line) with only 250 active queries (corresponding to a total of 750 samples in the model, since we still have the 500 initial samples from the source). The right panel of Fig.~\ref{fig:ALDAres} shows that the percentage of the support vectors from each domain with nonzero weights: in the target domain, this share increases constantly (\blu{solid} blue line), while it stabilizes for the samples from the source (to 40\% of the original training samples \blu{-- dashed red line}). This means that the importance of the source in the model is strongly reduced in the first iterations and then remains constant, while each new sample from the target becomes immediately important and receives a strong weight from the \blu{boosted} SVM classifier.

\begin{figure}
\begin{tabular}{cc}
\includegraphics[width=.43\linewidth]{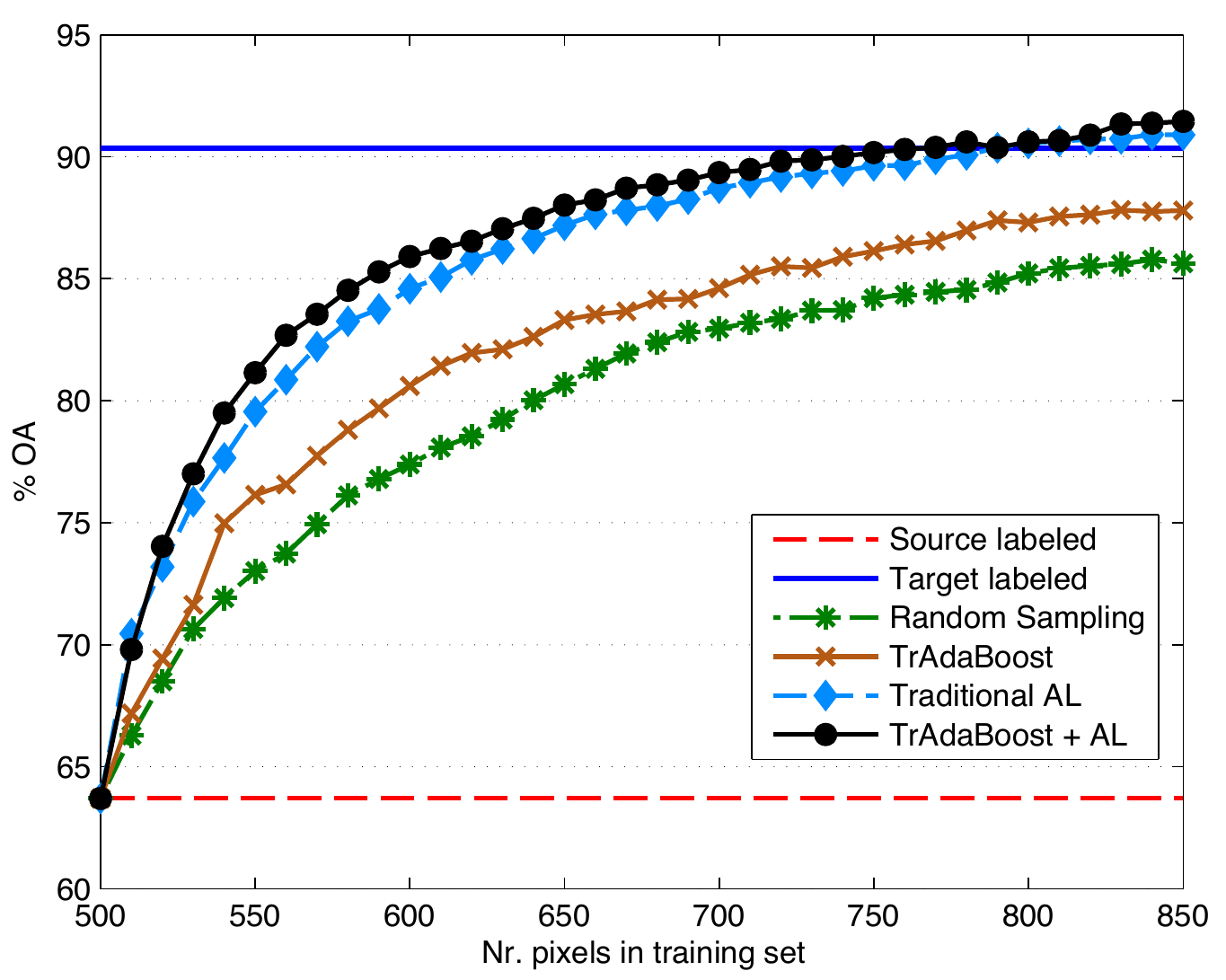}&
\includegraphics[width=.43\linewidth]{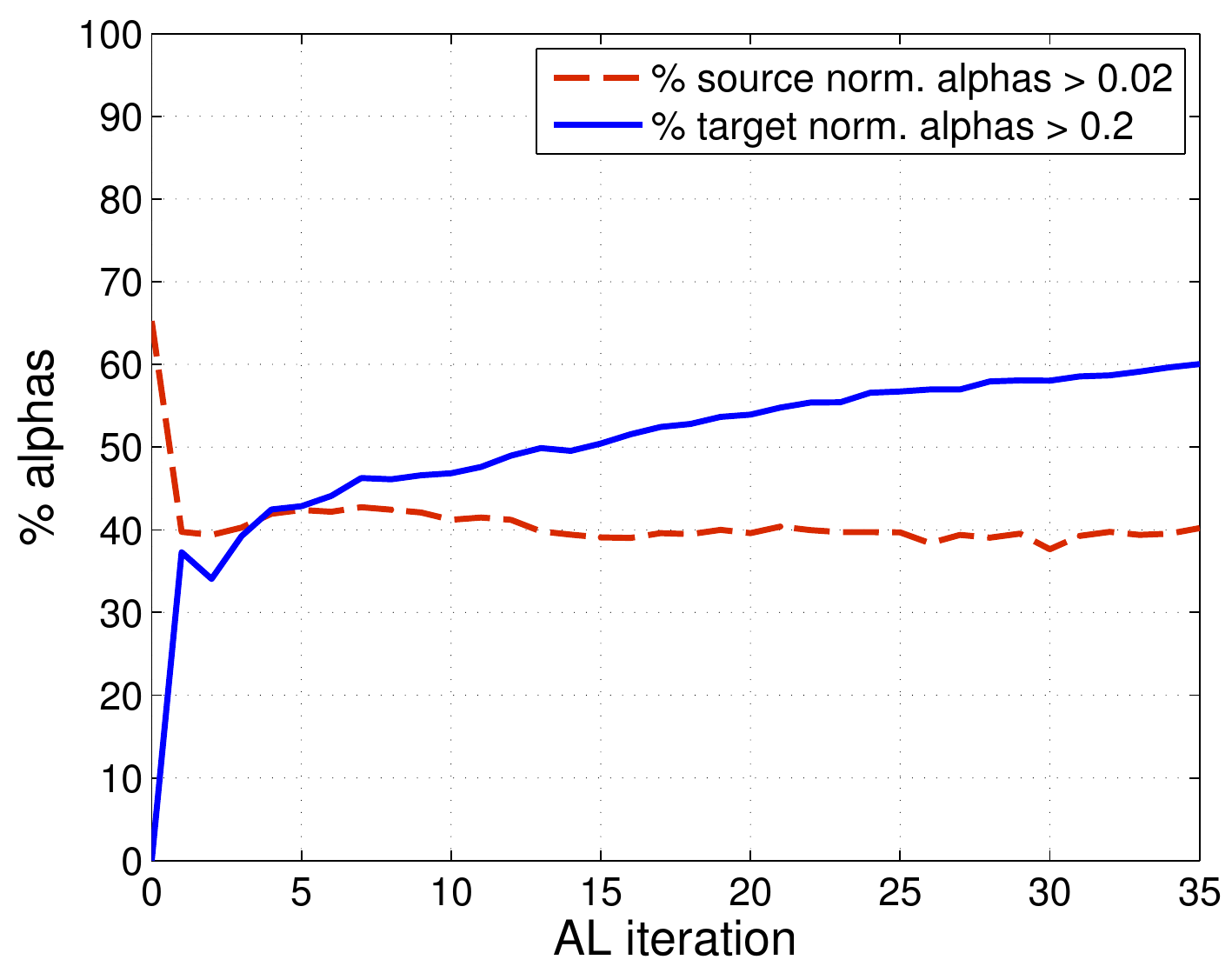}\\
\end{tabular}
\caption{AL results over the KSC data of Fig.~\ref{fig:KSC} (adapted from~\cite{Mat12}). Left: learning rates (Overall accuracy) for different methods. Right: percentage of source (dashed red line) and target (solid blue line) $\alpha$ weights larger than 0.02 (source) and 0.2 (target) along the iterative AL process in the `TrAdaBoost + AL' experiment (black line in the left panel).}
\label{fig:ALDAres}
\end{figure}

\section{\blu{Guidelines for the choice of the adaptation strategy}}
\blu{In this section, we will first provide guidelines for the selection of the most appropriate adaptation strategy and then discuss the issue of the validation of the adapted models.}

\subsection{How to chose the adaptation strategy}
\blu{In the previous sections, we presented different approaches to domain adaptation, grouped in four families. Depending on the problem considered, an analyst can favor one or the other. However, there are some guidelines that should be taken into account. They will depend mainly on the data available (for instance the sensors to be used) and on the effort already provided by the analyst (for instance, whether a classifier is already available, or if labels in the target domain are available or can be acquired easily):}

\begin{itemize}
\item \blu{If the data to be used are acquired by different sensors, they are associated with different feature spaces. In this case, only heterogeneous domain adaptation (i.e. methods that allow to align spaces of different dimensionality) should be considered. Accordinagly, multiview feature-representation-transfer methods such as (k)CCA or manifold alignment (see Section~\ref{sec:represent}) are the recommended choice.}

\item \blu{If a classifier trained on the source is already available -- and the effort of training is considerable -- methods of the third (adaptation of classifier) and fourth (adaptation by selective sampling) family should be preferred. These methods build on the model already defined on the source domain, while those of the two other families imply the definition of a new classifier that is successful in all domains.}

\item \blu{Whenever it is possible to acquire new labeled samples in the target, it should be done. There is no better way for correcting for a dataset shift than having examples of the class-conditional distribution in the target. Active learning and manifold alignment methods are to be preferred in that case.}

\item \blu{The level of dataset shift the methods can cope with goes along with their level of flexibility: representation methods relying on labeled samples from the target can cope with strong nonlinear deformations (since they allow for a kind of features registration between the domains), while those that do not use target samples (e.g., PCA, TCA, GM) are successful only if the data distributions are already pre-aligned and have not undergone drastic shifts (such as cases where the signature of a target class becomes identical of the one of another in the source). Among the unlabeled methods, the differences in their flexibility should be considered, going from linear global methods (e.g. PCA representation transfer) to local methods (e.g. based on clustering: GM, MA). If the first can address only rotations, translations and, to some extents, scalings of the data clouds, the others can model the per-sample transformation and allow more flexibility of the transform. The same type of reasoning holds for semisupervised methods: they will be able to correct for smaller shifts than methods based on active learning. When deployed in a domain adaptation setting, active learning mehtods collect target labeled samples that provide evidence of the real target class distributions, while the semisupervised method use only unlabeled data in the target and therefore cannot discover easily drastic changes in the distributions of the classes.}

\item \blu{The combination of methods of the different families is also possible. Selecting invariant features can be a preprocessing step to kernel manifold alignment, where the labels in the target domain have been acquired by active learning using the labels from the source domain.}

\end{itemize}

\blu{With these simple guidelines in mind, the analyst can select the most appropriate strategy (or combine a series of them) according to the considered data and application.}

\vspace{.5cm}

\subsection{How to validate}
\blu{A typical bottleneck for the employment of an adaptation strategy is the validation of the adaptation process itself, since it is assumed that no (or only few) labeled data are available for the target domain. Nonetheless, one should assess whether the adaptation was successful in the processing of the target image, even though no labeled samples are available for such validation. To address this crucial issue, a circular validation strategy is presented and applied to remote sensing images in~\cite{Bruzzone2010}. The strategy is based on the idea that an intrinsic structure relates the solutions consistent with the source and the target domains. A solution for the target domain (for which no prior information is available) is assumed to be consistent if the solution obtained by applying the same domain adaptation algorithm in the reverse sense (i.e., by using the classification labels in place of missing prior knowledge for target-domain instances) to source-domain data (considered as unlabeled in the reverse domain adaptation learning) is associated with an acceptable accuracy (which can be evaluated due to the available true labels for source-domain samples). This strategy can be effective for both understanding if the adaptation is feasible in the considered data set and selecting the most effective strategy.
}

\section{Conclusions}\label{sec:concl}
In this paper, we reviewed the recent advances in domain adaptation for remote sensing image analysis. Domain adaptation is a rising field of investigation in remote sensing, as it answers the need of re-using available ground reference samples to classify (or further process) new image acquisitions that may be covering different areas, at different time instants and possibly with different sensors. The increasing satellite data availability trend observed in the last years (in particular thanks to satellite constellations such as the Sentinels or the NASA A-train), as well as the commericalization of drone-mounted cameras pushed these problems at the forefront of researchers and analyst priorities.

We have reviewed the recent models proposed in the literature, grouped in four main families: i) the approaches based on the selection of invariant features, ii) those  based on the matching of the data representation, iii)  those based on the adaptation of the classifier trained on the source domain and iv) those based on limited, but effective sampling of labeled samples in the target domain. With practical examples, we have provided to the reader a thorough introduction to the field and some guidelines for the selection of the approaches to use in real application scenarios.

We believe that domain adaptation is of the highest importance to future Earth Observation, since multimodality and repeated imaging have become unavoidable \cite{Gom14}. The data are already there and new challenging problems can now be tackled with remote sensing. The discipline has succeeded in  entering many new sectors of society and it is now time to provide the tools to the users to perform a trustable monitoring that can be obtained in different sensor configuration or modalities. We think that domain adaptation and, more in general, machine learning can contribute to provide an answer to this call.

\bibliographystyle{IEEEbib}
\bibliography{GRSM-DA,LorenzoAddsOn}

\end{document}